%% file: main.tex
\documentclass[english]{cccconf}
\usepackage[comma,numbers,square,sort&compress]{natbib}
\usepackage{epstopdf}
\usepackage{times} % assumes new font selection scheme installed
\usepackage{amsmath} % assumes amsmath package installed
\usepackage{amssymb}  % assumes amsmath package installed
\usepackage{color}
\usepackage{booktabs}
\usepackage{graphicx}
\usepackage{subfigure}
\usepackage{multirow,booktabs}
\usepackage{stfloats} %fig location
\usepackage{fancyhdr}
\usepackage{footnote}
\usepackage{float}
\usepackage{pifont}
\usepackage[colorlinks, linkcolor=red, citecolor=blue]{hyperref}

\begin{document}

\title{Fast-SegSim: Real-Time Open-Vocabulary Segmentation \\ for Robotics in Simulation}

\author{Xuan Yu\aref{zju},
        Yuxuan Xie\aref{zju},
        Shichao Zhai\aref{zju},
        Shuhao Ye\aref{zju},
        Rong Xiong\aref{zju},
        Yue Wang\aref{zju}}

\affiliation[zju]{Department of Control Science and Engineering,
        Zhejiang University, Hangzhou, Zhejiang, P.~R.~China
        \email{ywang24@zju.edu.cn}}
% \affiliation[hit]{Harbin Institute of Technology, Harbin 150001, P.~R.~China
%         \email{xxx@hit.edu.cn}}

\maketitle

\input{sec0_abstract}

\input{sec1_intro}

\input{sec2_related}
\input{sec3_method}

\input{sec4_exp}
\input{sec5_conclusion}
\input{bibliography}

\end{document}

%% file: sec0_abstract.tex
\begin{abstract}
Open-vocabulary panoptic reconstruction is crucial for advanced robotics and simulation. However, existing $3\text{D}$ reconstruction methods, such as NeRF or Gaussian Splatting variants, often struggle to achieve the real-time inference frequency required by robotic control loops. Existing methods incur prohibitive latency when processing the high-dimensional features required for robust open-vocabulary segmentation.
We propose Fast-SegSim, a novel, simple, and end-to-end framework built upon $2\text{D}$ Gaussian Splatting, designed to realize real-time, high-fidelity, and $3\text{D}$-consistent open-vocabulary segmentation reconstruction. Our core contribution is a highly optimized rendering pipeline that specifically addresses the computational bottleneck of high-channel segmentation feature accumulation. We introduce two key optimizations: Precise Tile Intersection to reduce rasterization redundancy, and a novel Top-$K$ Hard Selection strategy. This strategy leverages the geometric sparsity inherent in the $2\text{D}$ Gaussian representation to greatly simplify feature accumulation and alleviate bandwidth limitations, achieving render rates exceeding $50$ FPS.
Fast-SegSim provides critical value in robotic applications: it serves both as a high-frequency sensor input for simulation platforms like Gazebo, and its $3\text{D}$-consistent outputs provide essential multi-view 'ground truth' labels for fine-tuning downstream perception tasks. We demonstrate this utility by using the generated labels to fine-tune the perception module in object goal navigation, successfully doubling the navigation success rate. Our superior rendering speed and practical utility underscore Fast-SegSim's potential to bridge the sim-to-real gap.
\end{abstract}


%% file: sec1_intro.tex
\section{INTRODUCTION}

% 1. 背景：开放世界的3D场景全景重建，机器人领域的应用价值。现有方法范式。
Open-world panoptic reconstruction stands as a cornerstone task in $3\text{D}$ scene understanding, providing the dense geometric and semantic information essential for embodied artificial intelligence and advanced robotics. Given the high costs associated with $3\text{D}$ data annotation and the powerful segmentation capabilities of recent $2\text{D}$ Vision-Language Models (VLMs) \cite{sam, ren2024grounded}, the prevailing paradigm in $3\text{D}$ panoptic reconstruction shifts towards leveraging $2\text{D}$ VLM outputs, such as masks or features, and lifting them into a $3\text{D}$ representation \cite{Chen2024PVLFF, yu2024panopticrecon, pr++, xie2025panopticsplatting}.

% 2. 现有方法（NeRF）：性能不足
Early methods in this domain are primarily based on Neural Radiance Fields (NeRF) \cite{lifting, pr++}. While works like Panoptic Lifting \cite{lifting} and PanopticRecon++ \cite{pr++} establish multi-view consistent labeling and end-to-end optimization strategies, NeRF’s implicit representation suffers from inherent limitations: intensive computation due to random sampling, and consequently, slow rendering speeds. Crucially, the latency incurred by NeRF-based rendering is often prohibitive for closing the high-frequency perception-action loop required by real-time robotic control and simulation platforms. Furthermore, NeRF’s implicit encoding makes it challenging to apply the learned representation directly to downstream tasks such as scene editing or simulation integration \cite{decomposingnerf}.

% 3. 现有方法（Gaussian Splatting）：新的性能瓶颈
The Gaussian Splatting family of methods emerges as an explicit representation that provides an order-of-magnitude increase in rendering speed and facilitates explicit $3\text{D}$ editing. Various works extend this framework for $3\text{D}$ scene understanding by augmenting Gaussians with feature attributes \cite{gaussiangrouping, opengaussian, langsplat}. However, most methods extending this explicit representation for panoptic segmentation are historically multi-staged, relying on error-prone prior processes for $2\text{D}$ mask alignment or feature distillation \cite{gaussiangrouping, PLGS}. While PanopticSplatting \cite{xie2025panopticsplatting} successfully introduces an end-to-end framework to address the staging issue, the reliance on rendering high-dimensional segmentation label images still leads to a significant drop in inference speed, making it insufficient for meeting the strict real-time requirements of robotics.

More importantly, despite the speed benefit of 3DGS for color rendering, these segmentation methods introduce a new bottleneck: rendering segmentation requires the accumulation of high-dimensional feature vectors (high channel count $C$) from all overlapping Gaussians ($N$) per pixel. This step drastically increases memory bandwidth usage and computation, making it impossible for these methods to maintain the real-time speed needed for robotic applications, especially when rendering segmentation results derived from high-dimensional features.

\begin{figure}[t]
    \centering
    \includegraphics[width=\linewidth,keepaspectratio]{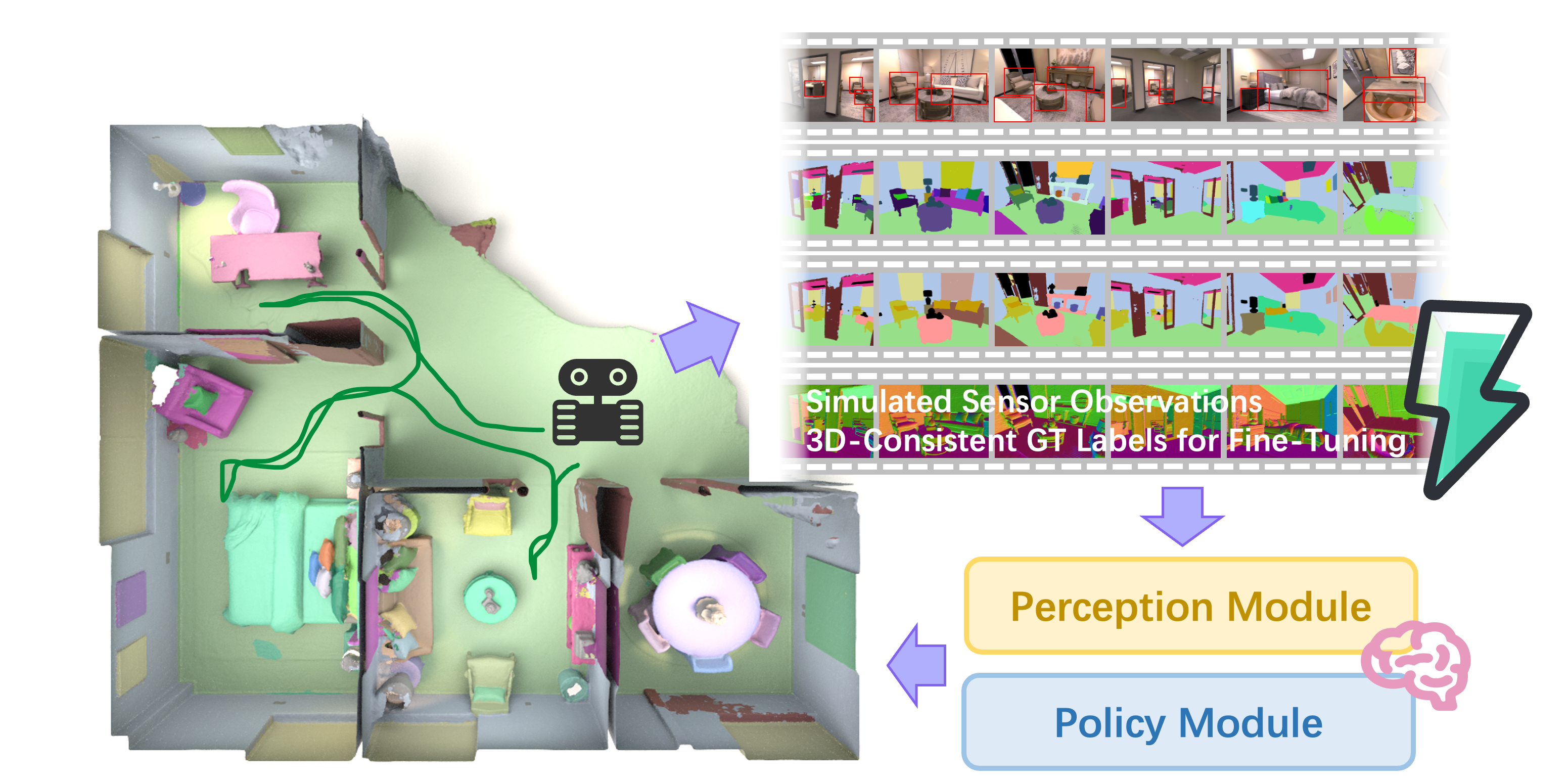}\\
    % \vspace{-0.3cm}
    \caption{Our real-time rendering speed enables robots to acquire realistic sensor observations and segmentation GT with low latency. The 3D-Consistent GT labels are critical for fine-tuning the Perception Module, effectively bridging the simulation and real-world gaps in embodied AI.}
    \vspace{-0.3cm}
    \label{fig:first_fig}
\end{figure}

% 4. 提出我们的方法和核心贡献（聚焦实时性）
To address this critical gap between high-fidelity $3\text{D}$ segmentation and real-time inference requirements, we introduce Fast-SegSim, a simple, end-to-end acceleration framework for $3\text{D}$ consistent open-vocabulary panoptic reconstruction. Unlike prior work focused solely on reconstruction quality or basic speedup, we comprehensively design Fast-SegSim to balance rendering efficiency, geometric precision, and the unique challenges posed by high-channel feature rendering.

We present two key ideas: a) We introduce Precise Tile Intersection, a mechanism that reduces redundant rasterization assignments and improves geometric efficiency by accurately calculating the bounding box of Gaussian projections.
b) We propose a novel Top-$K$ Hard Selection strategy. This method exploits the geometric depth coherence inherent in the $2\text{D}$ Gaussian representation to limit feature accumulation only to the few most relevant Gaussians ($K$), significantly reducing the feature processing workload and achieving render rates exceeding $50$ FPS.

% 5. 强调应用价值
The practical impact of Fast-SegSim is demonstrated across two key robotic application scenarios. Firstly, our system provides $3\text{D}$-consistent segmentation and geometry for a robotic simulation platform, overcoming the latency issues of previous methods. Secondly, the $3\text{D}$ consistency of our reconstructed scenes allows us to generate high-quality, multi-view 'ground truth' labels for training and fine-tuning downstream perception models. We showcase this utility by demonstrating that labels generated from our model successfully double the navigation success rate of an object goal navigation agent, underscoring Fast-SegSim's potential to bridge the sim-to-real gap and accelerate research in embodied AI.

% 6. 总结贡献
In summary, our main contributions are:

\begin{itemize}
 \item We propose Fast-SegSim, an end-to-end framework that achieves real-time, high-fidelity $3\text{D}$ open-vocabulary panoptic reconstruction by targeting the high-channel feature accumulation bottleneck.
 \item We introduce two core technical optimizations, Precise Tile Intersection and Top-$K$ Hard Selection, which significantly accelerate the rendering pipeline to achieve speeds over $50$ FPS.
 \item We validate the system's application value by demonstrating its use as a real-time sensor input for robotic simulation and as a generator of $3\text{D}$-consistent ground truth labels for boosting robot navigation performance.
\end{itemize}

%% file: sec2_related.tex
\section{RELATED WORKS}

\subsection{3D Panoptic and Open-Vocabulary Reconstruction}
\label{sec:related_panoptic_recon}

% 传统方法回顾 (Close-vocabulary, implicit/explicit)
Early efforts in $3\text{D}$ scene understanding often relied on explicit representations like occupancy grids or point clouds, adapting pre-trained $2\text{D}$ segmentation models to achieve semantic mapping \cite{rosinol2020kimera}. These methods, however, suffer from performance limitations dictated by the capacity of the $2\text{D}$ model and the robustness of the fusion strategy. More recently, studies leverage implicit neural representations (e.g., NeRF \cite{semanticnerf, pnf, instancenerf}) or explicit Gaussian representations \cite{zhu2024sni-slam} to encode $2\text{D}$ labels into a unified $3\text{D}$ feature space.

% NeRF-based Panoptic (Closed-set)
Within the implicit domain, methods closer to our work employ NeRFs to address $3\text{D}$ panoptic segmentation. Panoptic Lifting \cite{lifting} proposes a multi-view consistent label lifting scheme using linear assignment. PNF \cite{pnf} and Instance-NeRF \cite{instancenerf} attempt to obtain panoptic fields, though the latter's reliance on extracting discrete $3\text{D}$ masks for alignment is susceptible to scene variations. Many existing works \cite{wang2022dm, lifting, bhalgat2023contrastive} lift $2\text{D}$ instance masks to $3\text{D}$ using optimal supervision association or contrastive learning. However, these methods, predominantly closed-vocabulary, are limited by their confinement to predefined categories and the inherent slow rendering speed of NeRF’s implicit representation and volumetric ray-marching. This high computational cost and time requirement make them infeasible for real-time applications, such as robotic control loops.

% Open-Vocabulary Reconstruction (Feature Lifting)
To overcome the limitations of closed-vocabulary systems, a new wave of methods integrates $2\text{D}$ foundation models (like CLIP \cite{clip} and SAM \cite{sam}) to achieve open-vocabulary reconstruction. Feature-lifting methods, such as LERF \cite{kerr2023lerf} and DFF \cite{dff}, distill CLIP features into the $3\text{D}$ feature field, enabling flexible querying with text prompts. While effective for semantic querying, these methods often struggle to distinguish individual instances and result in rough segmentation boundaries.

\subsection{Gaussian Splatting for Panoptic Segmentation}
\label{sec:related_gs_panoptic}

% GS的优势和3D理解应用
The Gaussian Splatting family of methods (including both $3\text{D}$ and $2\text{D}$ GS) has gained rapid adoption due to its explicit representation, which offers an exceptional boost in rendering speed and geometric flexibility. Various works extend this framework for $3\text{D}$ scene understanding by augmenting Gaussians with feature attributes \cite{LEGaussian, langsplat, feature3dgs, gaussiangrouping, opengaussian}. For instance, LEGaussians \cite{LEGaussian} and LangSplat \cite{langsplat} introduce language semantics to Gaussians, while Feature 3DGS \cite{feature3dgs} distills high-dimensional features using a parallel rasterizer.

% 传统 GS Panoptic 的流程 (Mask-lifting & Multi-staged)
For the $3\text{D}$ panoptic reconstruction task, existing GS-based approaches generally fall into two categories:
a) Mask-Lifting Methods \cite{gaussian_grouping, PLGS}: These works utilize $2\text{D}$ segmentation masks (e.g., from Grounded SAM) to supervise $3\text{D}$ models. They face the critical challenge of misaligned $2\text{D}$ instance IDs across views and often rely on external tools like $2\text{D}$ trackers \cite{gaussian_grouping} or complex $3\text{D}$ matching \cite{PLGS} in a multi-stage pipeline, which can lead to error accumulation.
b) End-to-End Solutions: More recently, methods like PanopticSplatting \cite{xie2025panopticsplatting} introduce end-to-end frameworks to address the multi-staging issue.

% 强调新的瓶颈 (Critical Bottleneck)
Despite the success in improving reconstruction quality and achieving end-to-end training, a fundamental bottleneck persists across all methods that use feature-rich Gaussians for segmentation: rendering segmentation results derived from high-dimensional features. The reliance on accumulating high-channel feature vectors from numerous overlapping Gaussians dramatically increases the computational cost and memory bandwidth consumption. While PanopticSplatting offers an end-to-end solution, it still suffers from this high-channel rendering issue, leading to a significant drop in inference speed that fails to meet the stringent real-time requirements of robotic systems.

\subsection{Real-Time Rendering Acceleration}

% GS 加速的总体方向
The superior rendering speed of the Gaussian Splatting family of methods \cite{kerbl20233dgs, 2dgs, ligs} has inspired extensive research focused on further accelerating the representation and rendering pipeline. Current acceleration works primarily focus on reducing the model's complexity or optimizing the training process. For instance, FastGS \cite{ren2025fastgs} accelerates the training time of $3\text{D}$ GS, while Speedy-Splat \cite{hanson2025speedysplatfast3dgaussian} reduces the number of Gaussians through sparse primitives and pruning techniques to achieve faster rendering speeds and smaller model sizes. Other efforts have focused on adapting sampling strategies or improving the efficiency of the rasterization process itself.

% 现有加速工作的局限性
Crucially, these existing acceleration techniques are primarily designed and evaluated for low-channel output rendering, such as RGB (3 channels) or depth. They typically rely on reducing the total number of Gaussians ($N$) that contribute to the final image. Furthermore, most of these optimization techniques are tailored for the $3\text{D}$ Gaussian Splatting representation. As analyzed in the previous section, the primary bottleneck for $3\text{D}$ panoptic segmentation lies not just in the number of Gaussians, but in the massive cost of accumulating high-dimensional feature vectors (high channel count $C$). Existing acceleration methods, which mainly focus on $N$ reduction within the $3\text{D}$ GS paradigm, do not adequately address the unique rendering challenges presented by high-channel feature accumulation, nor are there specialized acceleration methods for the $2\text{D}$ Gaussian Splatting representation often preferred for its geometric quality.

% % 引入我们的工作 (Fast-SegSim)
% Our Fast-SegSim framework is the first to explicitly target this high-channel feature accumulation bottleneck. We do not rely on aggressive pruning or training acceleration. Instead, we introduce a novel rendering mechanism optimization that capitalizes on the geometric sparsity and depth coherence of the $2\text{D}$ Gaussian representation. By combining Precise Tile Intersection with the Top-$K$ Hard Selection strategy, we bypass the need to accumulate features from all overlapping Gaussians, resulting in a significant, architecture-level speedup that preserves high-fidelity segmentation quality while delivering the real-time frame rates demanded by robotic applications.

%% file: sec3_method.tex
\section{METHOD}
% 1.基于2D gaussian的特征表示
% Feature Representation Based on 3D Gaussian
% 为了表示场景的全景信息，我们在原始gs的基础上...，分别增加了语义和实例两个属性来分别学习类别level和物体level的特征
% 我们仿照rgb的blending过程，公式，具体而言，将gs的语义和实例特征投射到2d，...

% 2.FastGS
% 3DGS中更高效处理tile的方法
% 2DGS表征中需要注意的不同和近似处理

% 3.Instance Feature Field
% query-guided instance segmentation
% cuda中的具体设计

% 4. inference top K 
% 2DGS以及深度监督->几何好
% 机器人推理实时要求 ->top k方式

% 5.End-to-end Training
% 2D-3D Instance Assignment 2d标签关联
% total loss

\begin{figure*}[t]
    \centering
    \includegraphics[width=\linewidth,keepaspectratio]{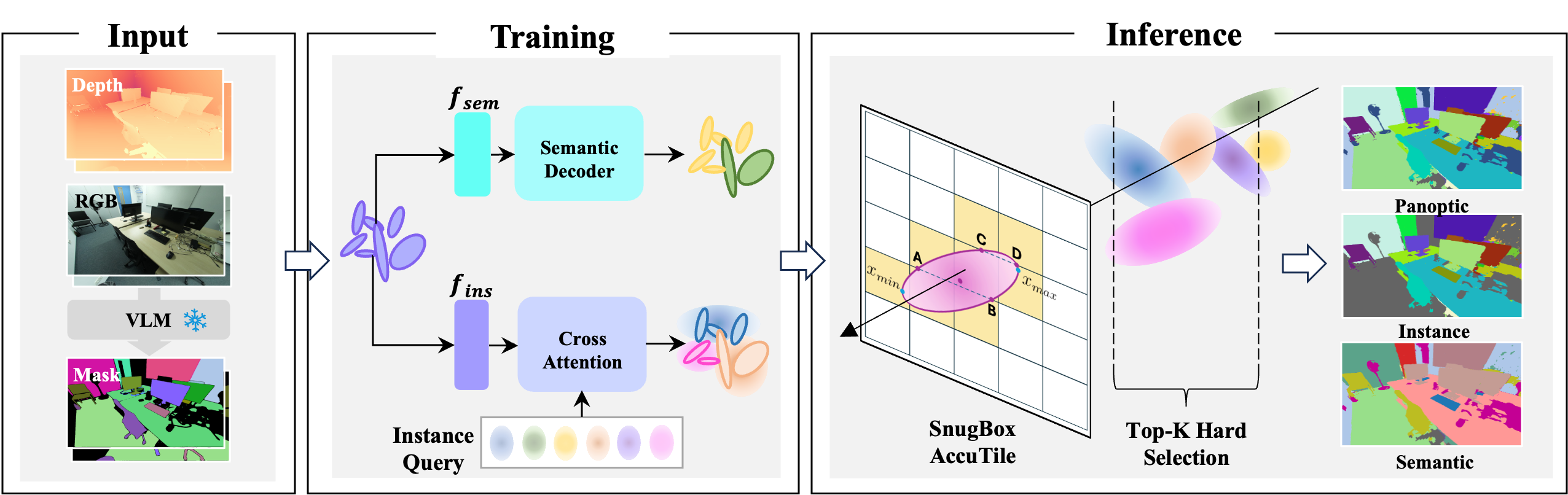}\\
    % \vspace{-0.3cm}
    \caption{The Input uses multi-view RGB-D frames alongside 2D supervision generated by VLM. During Training, the Gaussian Feature Field is learned, where the semantic feature ($\mathbf{f_{\text{sem}}}$) is processed by a Semantic Decoder, and the instance feature ($\mathbf{f_{\text{ins}}}$) is refined via Query-guided Cross Attention. For Inference, our highly optimized rendering pipeline is employed. The key to real-time performance is the Top-K Hard Selection strategy, which leverages the geometric sparsity inherent in Gaussian Splatting to significantly reduce the computational cost of high-dimensional feature accumulation within the SnugBox and AccuTile rasterization process, enabling real-time multi-modal rendering.}
    \vspace{-0.3cm}
    \label{fig:first_fig}
\end{figure*}

\subsection{Preliminaries}

2D Gaussian Splatting (2DGS) is a scene representation technique that models the environment as a collection of flattened Gaussian surfels. This approach is beneficial for high-fidelity surface reconstruction and excels at capturing thin structures. Each surfel $i$ is parameterized by its center $p_{i}\in\mathbb{R}^{3}$, 2D scale $(r_{u_{i}}, r_{v_{i}})$, and orientation vectors $t_{u_{i}},t_{v_{i}}\in\mathbb{R}^{3}$. The local tangent space allows a 2D point $\mathbf{u}=[u,v]^{\top}$ to be mapped to world coordinates $p(\mathbf{u})$ via:
\begin{equation}
    p(\mathbf{u})=p_{i}+r_{u_{i}}t_{u_{i}}u+r_{v_{i}}t_{v_{i}}v.
\end{equation}
The surfel is also defined by its Gaussian density $f(\mathbf{u})=\exp(-(u^{2}+v^{2})/2)$, an opacity $o_{i}$, and view-dependent color $c_{i}$. The surfel normal $n_i$ can be calculated from $t_{u_i}$ and $t_{v_i}$ by a cross-product, which is critical for geometric supervision. For rendering, surfels are depth-sorted and the final pixel color $C(\mathbf{x})$ is synthesized using accumulated opacity:
\begin{equation}
    C(\mathbf{x})=\sum_{i=1}^{N}c_{i}o_{i}f(\mathbf{u}_{i}(\mathbf{x}))\prod_{j=1}^{i-1}(1-o_{j}f(\mathbf{u}_{j}(\mathbf{x}))).
\end{equation}
Here, $\mathbf{u}_{i}(\mathbf{x})$ is the 2D projection of the pixel $\mathbf{x}$ onto the $i$-th surfel's tangent space. The efficiency of the accumulated blending makes 2DGS a highly suitable and differentiable representation for large-scale optimization.

%%%%%%%%%%%%%%%%%%%%%%%%%%%%%%
\subsection{Precise Tile Intersection for Surfels}

To mitigate the computational bottleneck of redundant tile assignments in 2DGS, we adapt the tile-based acceleration mechanisms, Snugbox and AccuTile, for our 2D Gaussian surfel representation. This adaptation requires robust and efficient computing of the projected inverse covariance matrix, $\mathbf{\Sigma}'^{-1}$, which defines the screen-space ellipse.

\textbf{Perspective-Corrected Conic Matrix Derivation.}
The core challenge lies in the complex projective transformation of the 2D {Perspective-Corrected Linearization scheme, which addresses the projection error inherent in simpler $2 \times 2$ Jacobian approximations while maintaining computational efficiency.

We first define the $3 \times 3$ homogeneous transformation matrix $\mathbf{T} \in \mathbb{R}^{3 \times 3}$, which maps a point $\mathbf{u} = [u, v, 1]^{\top}$ in the surfel's local tangent space to the homogeneous screen coordinates $\mathbf{x}' = [x', y', w']^{\top}$.
$$
\mathbf{x}' = \mathbf{T} \mathbf{u}
$$
The row vectors of $\mathbf{T}$ are denoted as $\mathbf{T}_{1}, \mathbf{T}_{2}$, and $\mathbf{T}_{w} \in \mathbb{R}^{3}$, incorporating the surfel's scale, 3D pose, and the camera's perspective projection. The Projected Covariance Matrix $\mathbf{\Sigma}' \in \mathbb{R}^{2 \times 2}$ is then approximated via a local linearization that explicitly includes the perspective term $\mathbf{T}_{w}$:
$$
\mathbf{\Sigma}' = \frac{1}{{\mathbf{T}_{w}}^2} \begin{pmatrix}
\mathbf{T}_{1} \cdot \mathbf{T}_{1} - \mathbf{T}_{w} \cdot \mathbf{T}_{1} & \mathbf{T}_{1} \cdot \mathbf{T}_{2} - \mathbf{T}_{w} \cdot \mathbf{T}_{2} \\
\mathbf{T}_{2} \cdot \mathbf{T}_{1} - \mathbf{T}_{w} \cdot \mathbf{T}_{2} & \mathbf{T}_{2} \cdot \mathbf{T}_{2} - \mathbf{T}_{w} \cdot \mathbf{T}_{2}
\end{pmatrix}
$$
The final $\mathbf{\Sigma}'^{-1} \in \mathbb{R}^{2 \times 2}$ required for AccuTile is the inverse of $\mathbf{\Sigma}'$:
$$
\mathbf{\Sigma}'^{-1} = (\mathbf{\Sigma}')^{-1} = \begin{pmatrix} A & B \\ B & C \end{pmatrix}
$$

\textbf{Snugbox and AccuTile Application.}
The computed $\mathbf{\Sigma}'^{-1}$ is passed to the AccuTile mechanism, which accurately determines the pixel extent $\mathbf{B}_{pixel}$ of the projected ellipse defined by the quadratic form $\mathbf{x}^{\top} \mathbf{\Sigma}'^{-1} \mathbf{x} \le k^2$, where $\mathbf{x} = [x, y]^{\top}$ are the image pixel coordinates and $\mathbf{k}$ is the surfel's density cutoff threshold. AccuTile achieves this by solving the intersection of the conic with horizontal and vertical lines, which simplifies to a standard quadratic equation, for example, in $x$:
$$
A x^2 + (2 B y_{c} + 2 D) x + (C y_{c}^2 + 2 E y_{c} + F) = 0
$$
The resulting tile bounding box $\mathbf{B}_{tile}$ is optimized by the Snugbox strategy. Snugbox minimizes redundant tile-to-surfel assignments and computational overhead by reducing outer loop iterations based on comparisons of tile spans along the X and Y axes ($x_{span}$ and $y_{span}$).
% $$
% \text{Minimize Loop Iterations } \propto \min(x_{span}, y_{span})
% $$
Snugbox's adaptive traversal path ensures the remaining assigned tiles are processed with minimal computational overhead, while AccuTile's high-precision boundary determination dramatically reduces the total number of redundant tile-to-surfel assignments, resulting in 2DGS rendering acceleration.

%%%%%%%%%%%%%%%%%%%%%%%%%%%%%%%%%%%%%%
% TODO
\subsection{Query-guided Instance Segmentation}

We address the challenge of misaligned instance IDs across multiple views, a crucial barrier to consistent scene-level instance segmentation, by lifting $2\text{D}$ instance masks to $3\text{D}$. To enforce $3\text{D}$ consistency and effectively associate these $2\text{D}$ labels, we leverage instance queries to explicitly guide the segmentation process within the $3\text{D}$ feature field.

A foundational design pattern in modern end-to-end segmentation frameworks involves using a set of object queries to govern detection and segmentation. In this architecture, queries interact with scene features to learn distinctive object characteristics, subsequently guiding the clustering of scene instances. The final segmentation of constituent scene units, such as $2\text{D}$ pixels or $3\text{D}$ primitives, is then determined by their affinity to these learned queries.

For $3\text{D}$ scene segmentation, previous studies \cite{pr++} model the affinity between $3\text{D}$ scene points and queries (represented as $3\text{D}$ Gaussian distributions) using a spatial distance-weighted attention mechanism. Adopting a similar principle, we define the segmentation of $3\text{D}$ Gaussians by utilizing distance-aware cross attention between the instance queries and the instance features associated with each Gaussian.

Given that the scene Gaussians are geometrically much smaller than the Gaussian distributions representing the queries, we treat the scene Gaussians as $3\text{D}$ points during the cross-attention computation. The similarity between a query feature $f_q$ and a Gaussian instance feature $f_{ins}$ is calculated as:
\begin{equation}
    {S}(f_q,f_{ins}) = {sigmoid}(f_q^T f_{ins})
\end{equation}

Furthermore, since our instance queries are characterized by a Gaussian distribution, the distance from a $3\text{D}$ point (the scene Gaussian's center) to the query can be naturally quantified by the query's probability density function. The spatial proximity weight $D$ between query $q$ (centered at $p_q$) and scene Gaussian $g$ (centered at $p_g$) is therefore defined based on their respective coordinates:
\begin{equation}
    D(p_q,p_g) = \frac{\varphi(p_g)}{\varphi(p_q)}
    \label{eq:wight_new}
\end{equation}
where $\varphi(\cdot)$ represents the probability density function of the query's Gaussian distribution. The total attention map $A$ is the combination of feature similarity and spatial proximity:
\begin{equation}
    {A}(q,g) = {S}(f_q,f_{ins}) {D}(p_q,p_g)
\label{gsours_new}
\end{equation}

The instance label prediction for a scene Gaussian $g$ is then derived by applying a softmax over its attention scores across all $N$ queries:
\begin{equation}
    l_{ins}(g)= softmax(\left[{A}(q_1,g) \ldots {A}(q_i,g) \ldots {A}(q_N,g) \right])
\end{equation}
Subsequently, the final instance label $I$ for a given pixel is generated by $\alpha$-blending the predicted Gaussian labels in a standard manner:
\begin{equation}
    I = \sum\limits_{i \in \mathcal{N}} l_{ins}(i) \alpha_i' \prod\limits_{j = 1}^{i-1} (1-\alpha_i')
    \label{eq:ins_blending_new}
\end{equation}

\subsection{Top-K Hard Selection for Real-Time Inference}

We capitalize on the geometric insight provided by the $2\text{D}$ Gaussian Splatting (2DGS) representation: objects are typically reconstructed as parsimonious, thin geometric surfaces. This inherent geometric sparsity suggests that representing accurate features, especially for dense prediction tasks like segmentation which tolerate less high-frequency noise than RGB, requires contributions from only a small, specific number of primitives.

In dense prediction tasks, the feature map channel dimensionality $C$ corresponds to the size of the feature vector (e.g., semantic features or instance embeddings), which is substantially higher than the 3 channels used for conventional RGB images. Given both a high $C$ and a potentially large set of overlapping Gaussians $N$ per pixel, the feature accumulation step represents a dominant computational and memory bandwidth bottleneck during rendering inference.

Based on this observation, we introduce the Top-$K$ Hard Selection mechanism to drastically optimize the feature accumulation stage. Rather than accumulating features from the entire set of $N$ overlapping Gaussians, we enforce a strict selection policy, utilizing only the top $K$ most contributory primitives based on a depth-aware criterion. This strategy fundamentally streamlines the inference pipeline.

By applying the Top-$K$ Hard Selection strategy, we restrict the feature summation to a fixed, limited set $\mathcal{K}$:
\begin{equation}
\label{eq:topk_acc_new}
\mathbf{F}_{\text{TopK}} = \sum_{i \in \mathcal{K}} \mathbf{c}_i \cdot \alpha_i \cdot \omega_i \quad \text{where } |\mathcal{K}| = K
\end{equation}
The set $\mathcal{K}$ is defined as the $K$ Gaussians, out of the total $N$, that contribute the highest cumulative influence after being sorted by visibility and depth. Since the total number of overlapping primitives $N$ is typically much greater than the small fixed threshold $K$ in dense scenes, this optimization substantially reduces the necessary feature memory access and computation required by the high channel count $C$ during feature accumulation, thereby achieving a marked acceleration in rendering inference speed with minimal impact on segmentation quality.

\subsection{End-to-end Panoptic Reconstruction}

After generating all necessary point-level predictions across the $3\text{D}$ scene, we derive the final pixel-level outputs, such as color, depth, and segmentation masks, through the standard $\alpha$-blending volume rendering process (refer to Eq.~\ref{vr}). The comprehensive supervision for the panoptic reconstruction is achieved by combining several loss terms, including geometric, photometric (appearance), semantic, and instance losses. The overall total loss $L$ is expressed as:
\begin{equation}
    L = L_{\text{depth}} + L_{\text{sdf}} + L_{\text{eik}} + L_{\text{ins}} + L_{\text{sem}} + L_{\text{pan}} + L_{c} + L_{\text{rgb}}
\end{equation}

{\bf Instance Query Alignment.}
To facilitate end-to-end supervision using $2\text{D}$ instance masks, which lack inter-view correspondence, a crucial step is the alignment between the model's predictions and the pseudo ground truth (GT) labels. We employ the Hungarian Algorithm for linear assignment, executed on a per-frame basis, to establish this critical connection.

Specifically, we first quantify the matching cost between the rendered binary instance masks and the pseudo GT binary masks for all possible pairs. Given that the number of instance queries is intentionally designed to be greater than the typical number of GT instances per frame, the Hungarian Algorithm finds the optimal assignment. This assignment minimizes the cumulative cost, thereby uniquely linking each GT instance to one of the predicted instance masks, providing the required consistent supervision.

\textbf{Loss Function Details.} The segmentation and rendering quality are optimized using the following specific loss formulations:

\textbf{Instance Loss.} Following the optimal instance assignment, the instance prediction branch is supervised using a combination of Dice loss and Binary Cross-Entropy (BCE) loss. This is calculated between the rendered instance mask $M_{\text{ins}}$ and its corresponding aligned GT mask $M_{\text{ins}}^{gt}$ for each matched instance:
\begin{equation}
\mathcal{L}_{\text{ins}} = \mathcal{L}_{\text{dice}}(M_{\text{ins}}, M_{\text{ins}}^{gt}) + \mathcal{L}_{\text{bce}}(M_{\text{ins}}, M_{\text{ins}}^{gt})
\label{eq:instance_loss_new}
\end{equation}

\textbf{Semantic Loss.} The semantic field is trained using a standard Cross-Entropy (CE) loss, computed between the rendered semantic map $M_{\text{sem}}$ and the provided semantic GT $M_{\text{sem}}^{gt}$:
\begin{equation}
    \mathcal{L}_{\text{sem}} = \mathcal{L}_{\text{ce}}(M_{\text{sem}}, M_{\text{sem}}^{gt})
\end{equation}

\textbf{Photometric Loss.} The accuracy of the novel view synthesis is enforced by supervising the rendered image $I$ against the ground-truth image $I^{gt}$. We use a blended loss that combines L1 distance with the Structural Similarity Index Measure (SSIM) to account for perceptual quality:
\begin{equation}
    \mathcal{L}_{\text{rgb}} = (1-\lambda_{\text{SSIM}})\mathcal{L}_1(I, I^{gt}) + \lambda_{\text{SSIM}}\mathcal{L}_{\text{SSIM}}(I, I^{gt})
\end{equation}
where $\lambda_{\text{SSIM}}$ is a weighting factor that balances the contributions of the two metrics.

%% file: sec4_exp.tex
\section{EXPERIMENTS}

Evaluation includes comparisons against state-of-the-art $3\text{D}$ segmentation models and quantitative analysis of our acceleration mechanisms. We showcase utility through two key robotic applications: Our $3\text{D}$-consistent segmentation "ground truth" effectively fine-tunes the object goal navigation perception model, significantly boosting success rate. We validate our geometry model's feasibility for real-time, multi-modal image rendering within a robotic simulation platform.

\subsection{Experimental Setup}

{\bf Datasets.} 
We utilize two established real-world indoor datasets for performance assessment: ScanNet-V2 \cite{dai2017scannet} and ScanNet++ \cite{yeshwanth2023scannet++}. ScanNet-V2 provides rich RGB-D streams, semantic labels, and reconstructed geometry, making it highly suitable for comprehensive $3\text{D}$ reconstruction and segmentation studies. ScanNet++ features high-resolution $3\text{D}$ scenes with detailed annotations, which is crucial for evaluating novel view synthesis quality and fine-grained $3\text{D}$ scene understanding. We select a representative subset of scenes, specifically $4$ scenes from ScanNet-V2 and $3$ scenes from ScanNet++, for our quantitative evaluations. Furthermore, we also validate the robotic navigation task across ten real-world indoor scenes captured using RGB and LiDAR, which are significantly larger in scale than the public datasets mentioned above.

{\bf Baselines.} 
We select a diverse group of representative baselines spanning both implicit (NeRF) and explicit (Gaussian) $3\text{D}$ panoptic segmentation approaches.
a) NeRF-based Methods: We compare against foundational implicit methods such as Panoptic Lifting \cite{lifting}, Contrastive Lift \cite{bhalgat2023contrastive}, PVLFF \cite{Chen2024PVLFF}, and PanopticRecon \cite{yu2024panopticrecon}, which exemplify $3\text{D}$ scene segmentation based on implicit neural representations.
b) Gaussian-based Methods: The explicit representation methods include Gaussian Grouping \cite{gaussiangrouping} and OpenGaussian \cite{opengaussian}, which lift $2\text{D}$ visual knowledge to the $3\text{D}$ Gaussian space for scene interpretation.

To ensure a fair comparison, we utilize the same $2\text{D}$ supervision masks generated by our VLM setup for all comparable methods, with exceptions made for PVLFF and Gaussian Grouping, which rely on their inherent mask generation protocols. For OpenGaussian, a post-processing step is added to convert its binary mask predictions into the required panoptic format.

{\bf Metrics.} 
In addition to standard $2\text{D}$ scene-level segmentation metrics, our evaluation rigorously focuses on real-time performance and practical application impact.
For Segmentation Quality, Following standard practices \cite{lifting}, we report Panoptic Quality (PQ), Semantic Quality (SQ), and Recognition Quality (RQ) for $2\text{D}$ panoptic segmentation, alongside mIoU and mAcc for semantic tasks, and mCov and mWCov for instance tasks.
For Inference Speed and Efficiency, we quantify the real-time capability by measuring the Rendering Time (in milliseconds and FPS), total  number of  Gaussians Processed (RN-Total) and the average number of Gaussians Processed per tile (RN/Tile) during inference, which validates the efficiency of our acceleration module.
For Application Performance, we directly evaluate the practical impact by measuring the Navigation Success Rate (SR) of a robotic agent, specifically to quantify the performance gain achieved by fine-tuning its perception module with the $3\text{D}$-consistent segmentation "ground truth" labels generated by our method in object goal navigation tasks.

{\bf Implementation Details.}
We employ a $2\text{D}$ foundation model, specifically Grounded-SAM \cite{ren2024grounded}, to generate the initial $2\text{D}$ semantic and instance masks used as supervision. Our comparative and ablation studies are anchored on a robust Gaussian base model, $\text{LI-GS}$ \cite{ligs}, which is built on Gaussian surfels for scene reconstruction. We configure the feature vector dimensions such that the Gaussian semantic features are $16$-dimensional, while the instance features for both Gaussians and queries are set to $32$-dimensional.

\begin{table*}[]
\caption{Panoptic Segmentation quality using different methods}
\centering
\setlength{\tabcolsep}{5pt}
\renewcommand\arraystretch{1.2}
\resizebox{\linewidth}{!}{
\begin{tabular}{lccc|cc|cc|ccc|cc|cc}
\toprule
\multirow{2}{*}{\textbf{Method}} & \multicolumn{7}{c}{ScanNet} & \multicolumn{7}{c}{ScanNet++} \\
\cmidrule(r){2-8} \cmidrule(r){9-15}
 & PQ $\uparrow$ & SQ $\uparrow$  & RQ $\uparrow$ & mIoU $\uparrow$ & mAcc $\uparrow$ & mCov $\uparrow$ & mW-Cov $\uparrow$ & PQ $\uparrow$ & SQ $\uparrow$ & RQ $\uparrow$ & mIoU $\uparrow$ & mAcc $\uparrow$ & mCov $\uparrow$ & mW-Cov $\uparrow$ \\
\midrule
Panoptic Lifting 
    & 57.86 & 61.96 & 85.31 & 67.91 & 78.59 & 45.88 & 59.93 
    & 71.14 & 77.48 & 88.14 & 81.34 & \textbf{89.67} & 56.17 & 68.51 \\
Contrastive Lift 
    & 37.35 & 41.91 & 57.60 & 64.77 & 75.80 & 13.21 & 23.26
    & 47.58 & 57.23 & 65.81 & 81.09 & 89.30 & 27.39 & 36.51 \\
PVLFF 
    & 30.11 & 51.71 & 44.43 & 55.41 & 63.96 & 45.75 & 48.41
    & 52.24 & 66.86 & 65.56 & 62.53 & 70.31 & 67.95 & 75.47 \\
% \midrule
Gaussian Grouping
    & 43.75 & 50.63 & 72.68 & 58.05 & 68.68 & 52.70 & 58.10
    & 33.10 & 40.60 & 67.27 & 59.53 & 68.13 & 29.83 & 36.83\\
OpenGaussian 
    & 48.73 & 51.48 & 88.10 & 54.05 & 68.43 & 44.43 & 49.60 
    & 51.03 & 56.93 & 85.73 & 61.80 & 73.97 & 50.00 & 51.02  \\
PanopticRecon
    & 63.70 & 64.81 & 81.17 & 68.62 & 80.87 & 66.58 & 77.84
    & 68.29 & 77.01 & 85.05 & 77.75 & 87.08 & 51.34 & 62.79 \\
\textbf{Ours} 
    % scannet
    & \textbf{74.12} 
    & \textbf{74.12}
    & \textbf{100.0}
    & \textbf{74.93}
    & \textbf{83.07}
    & \textbf{72.89}
    & \textbf{79.14}
    % scannet++
    & \textbf{77.09}
    & \textbf{82.36}
    & \textbf{93.60}
    & \textbf{81.77}
    & {89.01}
    & \textbf{74.23}
    & \textbf{77.91}\\
\bottomrule
\end{tabular}
}
\label{tab:Comparative_baseline}
\vspace{-0.3cm}
\end{table*}

\begin{figure}[t]
    \centering
    \includegraphics[width=\linewidth,keepaspectratio]{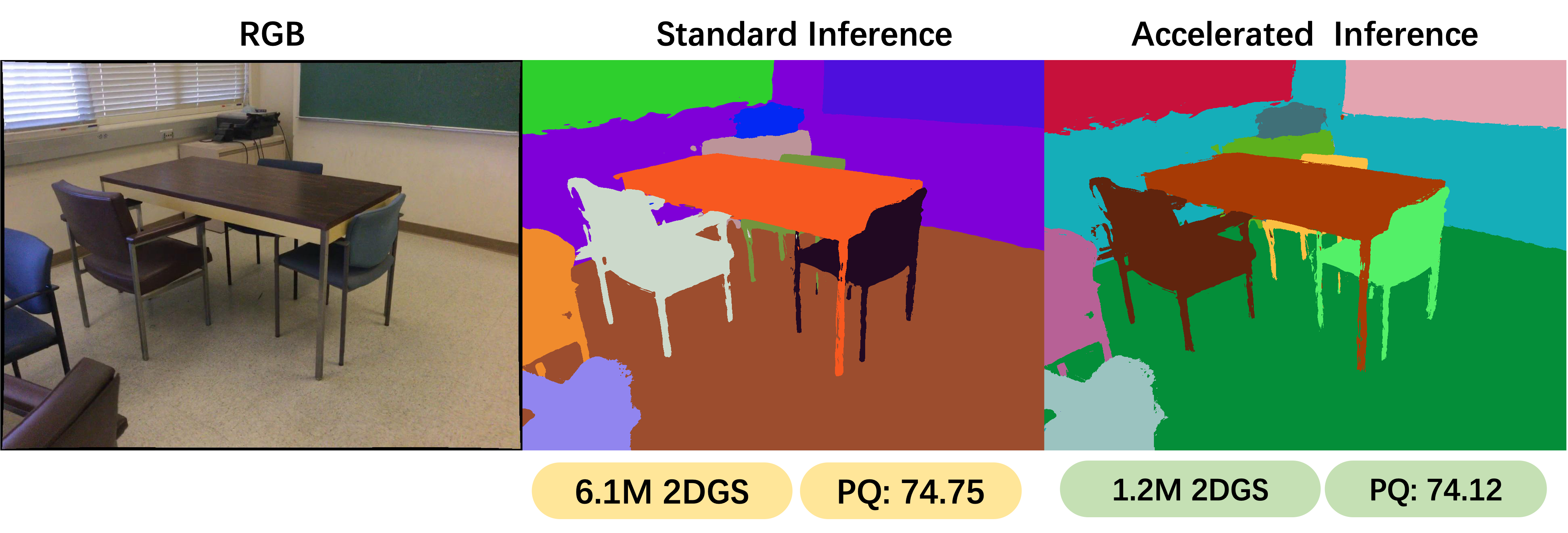}\\
    % \vspace{-0.3cm}
    \caption{Comparison of Fast-SegSim before and after acceleration: while significantly reducing the number of rendered Gaussians and substantially increasing rendering speed, segmentation quality remains highly consistent.}
    \vspace{-0.3cm}
    \label{fig:exp2}
\end{figure}

\subsection{Comparative Study}
Quantitative results comparing our approach with both NeRF-based and Gaussian-based baselines across the ScanNet-V2 and ScanNet++ datasets are presented in Tab.~\ref{tab:Comparative_baseline}.

As evidenced by the metrics, our proposed method significantly surpasses both NeRF-based and Gaussian-based counterparts. We proceed by analyzing the primary sources of segmentation errors in the comparative approaches, which consequently substantiates the robustness and efficacy of our methodology.

% TODO
First, multi-stage approaches such as PanopticRecon and Gaussian Grouping are prone to error propagation due to their pipeline design. Early-stage inaccuracies—such as coarse or incomplete 2D segmentation—tend to compound in later stages, ultimately degrading the quality of the final 3D panoptic output. In contrast, our method adopts a unified end-to-end framework that jointly optimizes geometry, semantics, and instance representation, thereby avoiding cascaded errors and enabling globally consistent reconstruction.

Another limitation observed in several baselines—particularly PanopticLifting and OpenGaussian—is the lack of consistent instance labeling across views. These methods often assign different identifiers to the same physical object when observed from varying viewpoints, which significantly harms scene-level evaluation metrics. Our approach addresses this by explicitly representing each 3D instance with a dedicated learnable query token. This design jointly encodes both appearance features and 3D spatial priors, ensuring that multi-view observations of the same object are consistently mapped to a single, coherent instance identity.

Moreover, feature-lifting strategies like Contrastive Lift encounter difficulties in disambiguating objects with visually similar appearances, leading to merged or missing segments. While OpenGaussian attempts to incorporate spatial structure through discretization and clustering, the resulting spatial priors remain relatively coarse and insufficient for fine-grained separation. In comparison, our model employs deformable, learnable instance queries together with a distance-aware similarity measure. This mechanism enables precise spatial reasoning during segmentation, effectively distinguishing adjacent objects even when their visual features are highly similar.

\subsection{Ablation Study}

To validate the effectiveness of our two key contributions, Precise Tile Intersection and Top-$K$ Hard Selection, in accelerating real-time rendering inference, we conduct a detailed ablation study. In robotic simulation scenarios, real-time rendering of segmented images from the camera's perspective is critical for high-response and low-latency robot operations. Our experiments aim to quantify how these components significantly boost the rendering speed, thereby ensuring real-time segmented image rendering in our reconstructed simulation scenes.

Table \ref{tab:ablation_speed} presents the rendering performance under different configurations, focusing primarily on inference time, frame rate, and key efficiency metrics corresponding to the two optimization mechanisms.

\begin{table}[t]
\centering
\caption{Impact of Precise Tile Intersection and Top-$K$ Hard Selection on Rendering Inference Speed.}
\label{tab:ablation_speed}
    \begin{tabular}{lccccc}
    \toprule
    \textbf{Exp} & \textbf{Time} $\downarrow$ & \textbf{FPS} $\uparrow$ & \textbf{RN-Total} $\downarrow$ & \textbf{RN / Tile} $\downarrow$ & \textbf{PQ} $\uparrow$ \\
    \midrule
        (A)  & 46 & 22 & 6.1M & 1230 & 74.75 \\ %Baseline (w/o AccuTile/Top-K)
        (B)  & 40 & 25 & 3.9M & 800 & 74.73 \\ %Only Precise Tile Intersection (w/ AccuTile)
        (C)  & 27 & 37 & 1.2M & 610 & 74.17 \\ %Only Top-$K$ Hard Selection ($K=3$)
        (D)  & 22 & 45 & 1.2M & 580 & 74.12 \\ %Full Method (AccuTile + Top-$K$, $K=3$)
    \bottomrule
    \end{tabular}
    \vspace{-0.3cm}
\end{table}

\textbf{Precise Tile Intersection. }
Configuration (B) is designed to evaluate the effectiveness of the Precise Tile Intersection mechanism (implemented via Snugbox and AccuTile) introduced. This method significantly reduces redundant Tile-to-Surfel assignments by approximately computing the inverse covariance matrix of the Gaussian projection, $\Sigma^{\prime-1}$, thereby lowering the computational overhead of the rasterization stage. As shown in Table \ref{tab:ablation_speed}, switching from configuration (A) to (B) reduces the Average Tile Allocation Count from 6.1 M to 3.9 M. This demonstrates that this mechanism effectively mitigates the bottleneck of the Tile assignment stage and improves the rendering efficiency of 2DGS.

\textbf{Top-$K$ Hard Selection. }
Configuration (C) validates the Top-$K$ Hard Selection strategy proposed. This strategy aims to address the feature accumulation bottleneck caused by the high channel count $C$ in semantic/instance/panoptic segmentation tasks. By leveraging the geometric advantage of the 2DGS representation, we cease feature accumulation for all $N$ overlapping Gaussians, selecting instead only the top $K$ most contributory Gaussians based on the depth criterion. By limiting the feature accumulation to $K=24$, we reduce the computational complexity of feature accumulation.

As Table \ref{tab:ablation_speed} shows, configuration (C), compared to the baseline (A), sees a drastic reduction in rendered the total count of surfels from 6.1 M to 1.2 M, with inference time dropping to 27 ms. This proves the superior acceleration effect of the Top-$K$ strategy when dealing with high-dimensional features, achieving accelerated rendering inference without significantly sacrificing segmentation quality.

Finally, the Full Method (D) combines the benefits of both Precise Tile Intersection and Top-$K$ Hard Selection, optimizing both the rasterization and feature accumulation bottlenecks. This achieves an inference time of 22 ms and a frame rate of up to 45 FPS, which is satisfied the requirements for real-time robotic applications.

\subsection{User Case}

To validate the practical application and value of our real-time segmentation rendering approach in the robotics domain, we conduct experiments across two distinct use cases: enhancing navigation task perception and enabling real-time rendering simulation.

\begin{table}[t]
\centering
\vspace{-0.3cm}
\caption{Impact of finetuning with 3D-consistent label ground-truth from Fast-SegSim on navigation task.}
\label{tab:nav}
    \begin{tabular}{lcc}
    \toprule
    \textbf{Scene} & \textbf{SR}-Before & \textbf{SR}-After  \\
    \midrule
        Laboratory Room  & 20\% & 100\%\\ %Baseline (w/o AccuTile/Top-K)
        House  & 25\% & 50\% \\ %Only Precise Tile Intersection (w/ AccuTile)
    \bottomrule
    \end{tabular}
    \vspace{-0.3cm}
\end{table}

{\bf Navigation Task.}
Our primary contribution to robotics is leveraging the 3D-consistent segmentation reconstruction model as 'ground truth' data to fine-tune the perception capability of downstream navigation tasks, thereby significantly boosting navigation success rates. 
% The core challenge in robot object goal navigation lies in maintaining accurate object perception, which is often limited by the inherent constraints of a single-frame observation. This limited visual field leads to a keyhole effect, where the robot's perception of the object and its context is incomplete and prone to errors.
We utilize RGB-D data collected from 10 similar indoor real-world scenes. For the navigation task, we employ the two-stage Object Goal Navigation algorithm based on VLFM \cite{vlfm}, which comprises a Perception Module and a Decision Module. The Perception Module initially uses a combination of a closed-set $2\text{D}$ detector~\cite{lei2025yolov13realtimeobjectdetection}, and an open-set $2\text{D}$ detector~\cite{liu2023grounding} integrated with a segmentation model~\cite{sam}.

As shown in Tab.~\ref{tab:nav}, when tested directly on reconstructions of real-world data, the perception module frequently suffers from missed or false detections due to limited single-frame visibility, leading to a low navigation success rate.

To address this, we use the $3\text{D}$-consistent outputs of our segmentation reconstruction model as synthetic 'ground truth' labels in the form of $2\text{D}$ bounding boxes. We then fine-tune the closed-set detector within the Perception Module using these high-fidelity labels generated from our reconstructed scenes. Since our segmentation reconstruction model's outputs are inherently $3\text{D}$-consistent labels that reinforce common object categories across various viewpoints, this process effectively transfers robust, multi-view scene knowledge to the detector. This fine-tuning process dramatically improves the accuracy of the Perception Module in testing scenarios, resulting in a substantial increase in the overall navigation success rate.

\begin{figure}[t]
    \centering
    \includegraphics[width=\linewidth,keepaspectratio]{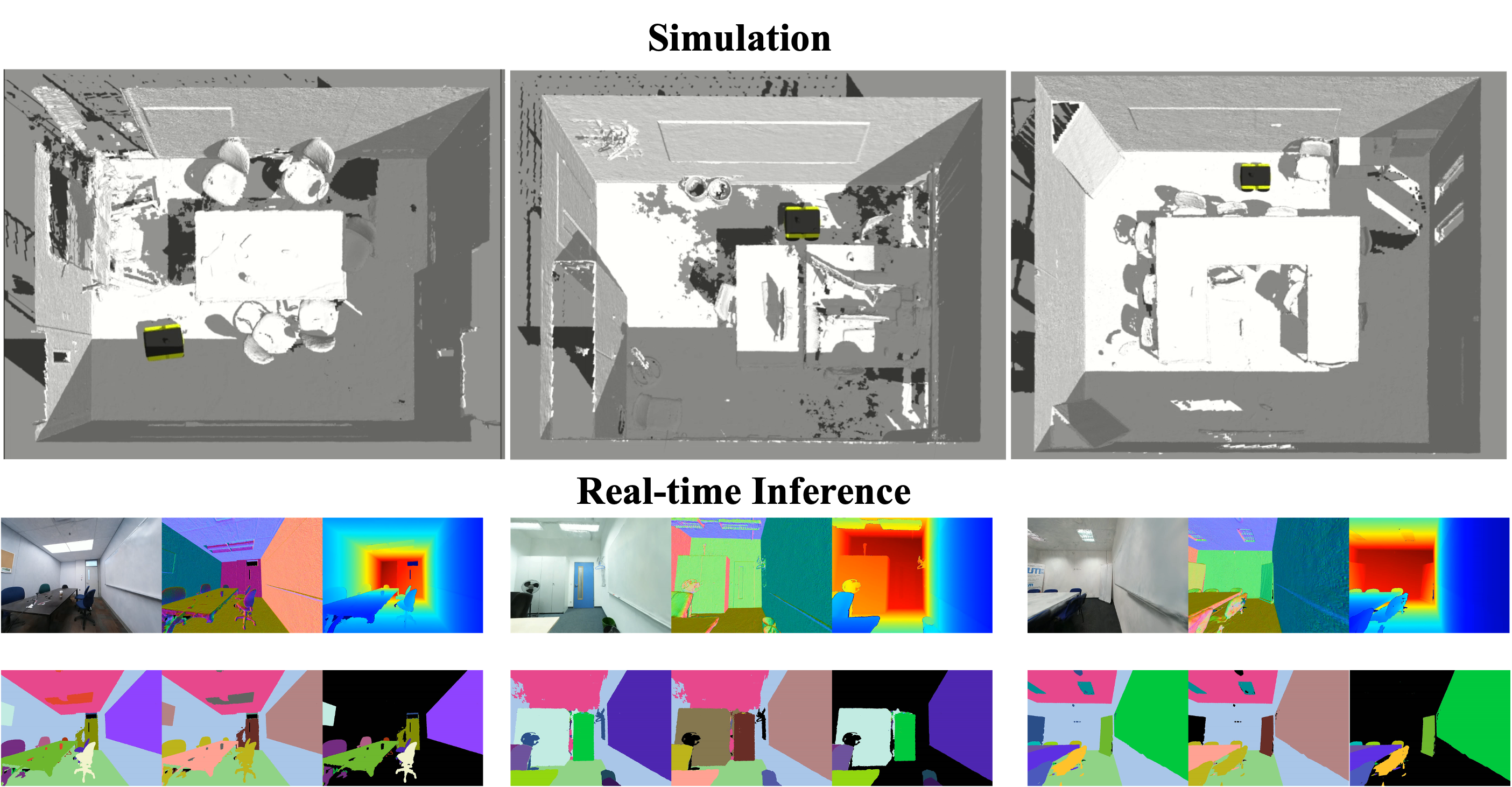}\\
    % \vspace{-0.3cm}
    \caption{Real-time Sensor Simulation with Fast-SegSim.}
    \vspace{-0.5cm}
    \label{fig:exp3}
\end{figure}

{\bf Real-time Rendering Simulation.}
We demonstrate the viability of our system for high-fidelity robot simulation. To facilitate realistic physical interactions between the robotic agent and the environment, we incorporate the meshes generated from models trained on three scenes of ScanNet++ into the Gazebo environment. Upon receiving navigation commands within Gazebo, the robot, exemplified by the Jackal UGV, initiates traversal.

During the robot's movement, the Gazebo simulation engine streams the robot's camera poses. Utilizing these poses, our system, Fast-SegSim, performs novel view synthesis to instantaneously generate photorealistic RGB, depth, and panoptic segmentation images. These rendered views serve as the essential sensor inputs for activating and testing downstream algorithms within the simulation framework. Critically, our inference process operates at a real-time frequency, reliably meeting the strict latency and high update rate demands of modern robotic simulation platforms. This ensured responsiveness is vital for closing the perception-action loop in dynamic environments.

%% file: sec5_conclusion.tex
\section{Conclusion}
We propose Fast-SegSim, a real-time, end-to-end framework for open-vocabulary panoptic reconstruction, featuring Precise Tile Intersection and Top-K Hard Selection to overcome high-dimensional rendering bottlenecks. It delivers high-fidelity, 3D-consistent segmentation at over 45 FPS. Fast-SegSim functions as a low-latency sensor simulator in Gazebo and generates multi-view “ground truth” labels that significantly improve downstream perception, bridging high-quality reconstruction and real-time robotics needs.

%% file: bibliography.tex
%\bibliographystyle{gbt7714-2005}
%\bibliography{bibliography}
\bibliographystyle{IEEEtran}
\bibliography{IEEEabrv,bibliography}

% \addcontentsline{toc}{chapter}{\bibname}